\documentclass{article}
\usepackage{spconf,amsmath,graphicx}
\usepackage{hyperref}
\usepackage{mathptmx} 
\usepackage[T1]{fontenc} 
\usepackage{lmodern} 
\usepackage{enumitem}
\setlist{nosep, leftmargin=14pt}
\usepackage{graphicx}
\usepackage{mwe} 
\usepackage{amsmath}
\usepackage{amssymb}
\usepackage{graphicx}
\usepackage{overpic}
\usepackage{tabularx}
\usepackage{booktabs} 
\usepackage{graphicx}  
\usepackage{caption}   
\usepackage{float}  
  
\title{Leveraging Gait Patterns as Biomarkers: An attention-guided Deep Multiple Instance Learning Network for Scoliosis Classification}

\name{Haiqing Li, Yuzhi Guo, Feng Jiang, Qifeng Zhou, Hehuan Ma, Junzhou Huang$^{\star}$ 
}

\address{Department of Computer Science and Engineering, The University of Texas at Arlington, Texas, USA}

\begin{document}
%
\maketitle
\begin{abstract}
Scoliosis is a spinal curvature disorder that is difficult to detect early and can compress the chest cavity, impacting respiratory function and cardiac health. 
Especially for adolescents, delayed detection and treatment result in worsening compression. 
Traditional scoliosis detection methods heavily rely on clinical expertise, and X-ray imaging poses radiation risks, limiting large-scale early screening. 
We propose an Attention-Guided Deep Multi-Instance Learning method (Gait-MIL) to effectively capture discriminative features from gait patterns, which is inspired by ScoNet-MT's pioneering use of gait patterns for scoliosis detection.  
We evaluate our method on the first large-scale dataset based on gait patterns for scoliosis classification. The results demonstrate that our study improves the performance of using gait as a biomarker for scoliosis detection, significantly enhances detection accuracy for the particularly challenging Neutral cases, where subtle indicators are often overlooked. 
Our Gait-MIL also performs robustly in imbalanced scenarios, making it a promising tool for large-scale scoliosis screening.
\end{abstract}
\section{Introduction}
\label{sec:intro}
\begin{figure}[t]  
    \centering
    \begin{overpic}[width=0.8\linewidth]{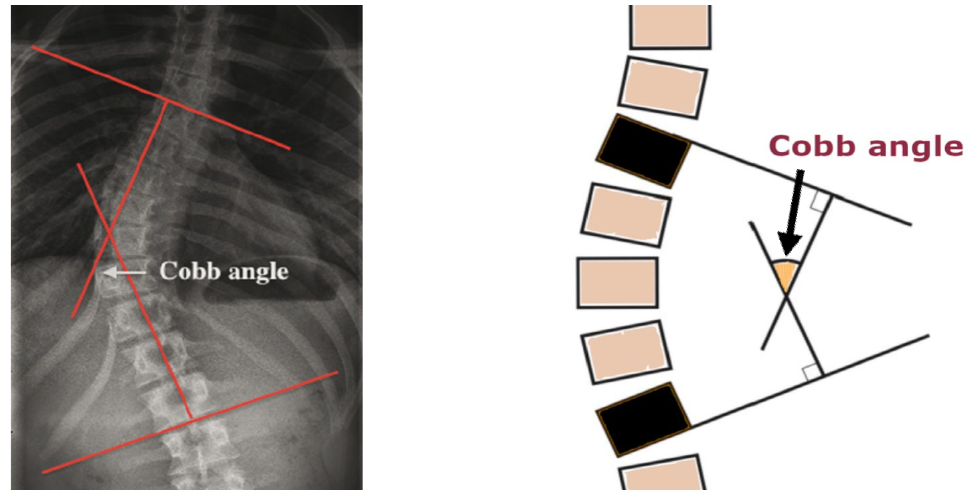}  
        \put(16,-3){(a)} 
        \put(75,-3){(b)}       
    \end{overpic}
    \caption{Illustrations of Cobb angle calculation, used to evaluate scoliosis severity.}
    \vspace{-0.5cm}
    \label{fig:cobb}
\end{figure}

Scoliosis is a spinal curvature disorder that is difficult to detect in its early stages, and for most children and adolescents, the cause is idiopathic~\cite{janicki2007scoliosis,goldberg2008scoliosis}. Without early detection and timely treatment, scoliosis can progressively worsen, not only affecting the posture but also potentially causing long-term back pain, breathing difficulties, cardiovascular dysfunction, and other issues, severely impacting their health and quality of life. 
Its diagnostic criteria primarily are assessing the degree of spinal curvature (usually measured by the Cobb angle on an X-ray, as shown in Fig.~\ref{fig:cobb} and the presence of asymmetry in the thorax, shoulders, or waist height~\cite{kotwicki2008evaluation}. If the curvature exceeds 10 degrees, scoliosis is diagnosed~\cite{weinstein2008adolescent}. This method not only requires extensive clinical experience but also exposes patients to radiation, which limits the feasibility of large-scale early screening. With the development of deep learning~\cite{li2023sclera}, researchers have started exploring non-contact and more efficient detection methods, such as back photo analysis~\cite{yang2019development}. These methods~\cite{kokabu2021algorithm,alharbi2020deep} have advanced the application and development of computer vision technology in the medical field, but they still face challenges related to privacy issues and the need for high user cooperation~\cite{dang2022avet}.

\begin{figure*}[t]  
    \centering
    \includegraphics[width=1\linewidth]{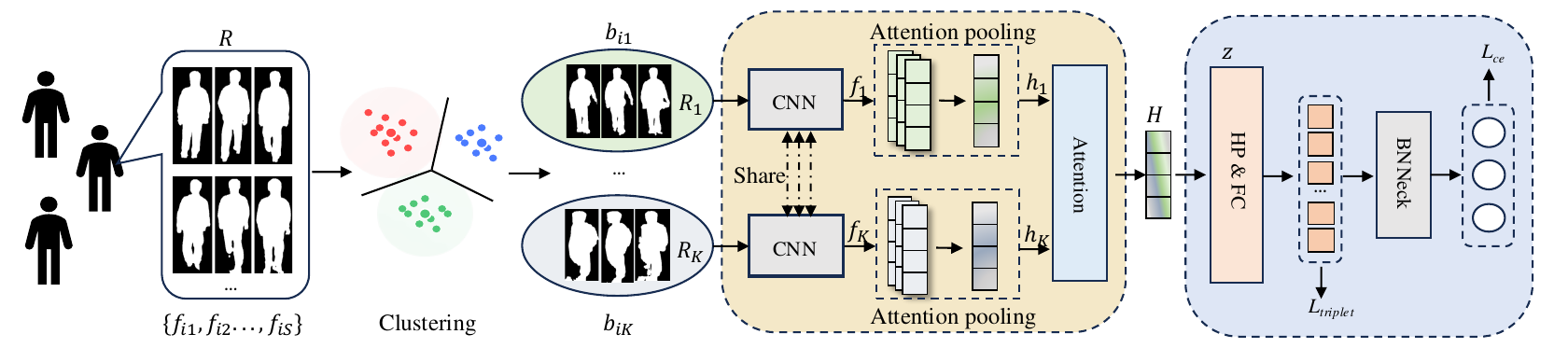}  
    \caption{The overall flowchart of the proposed Gait-MIL. It consists of four parts: (1) K-Means is used to cluster the sampled gait frames into distinct bags, enabling the model to extract specific features from different gait phases, (2) Attention-based pooling is applied to focus on the most informative frames within each bag, and (3) An attention mechanism is used to aggregate bag features into a global representation, (4) The triplet and cross-entropy losses are utilized to drive the training process. }
    \vspace{-0.5cm}
    \label{fig:framework}
\end{figure*}

To address this challenge, ScoNet-MT~\cite{zhou2024gait} first leverages gait patterns as biomarkers for scoliosis detection and introduce the first large-scale dataset based on gait patterns (Scoliosis1K) for scoliosis classification. ScoNet-MT processes each frame of participant videos by feeding individual silhouette images into an image encoder. The representations of all frames are then pooled and fed into a scoliosis predictor for final classification. However, gait patterns are inherently complex and dynamic, as each sequence consists of multiple frames, and not all frames contribute equally to the final classification. Multiple Instance Learning (MIL)~\cite{maron1997framework, ilse2018attention} has been demonstrated as a favorable option for addressing this issue, especially in situations where only some instances (frames) within a set (sequence) are informative for making a decision. 
 

Inspired by prior works~\cite{zhou2024gait,fan2023opengait}, we propose an Attention-Guided Deep Multiple Instance Learning Network (Gait-MIL) for scoliosis classification. Specifically, we cluster all frames in a sequence into several categories, which may represent different phases of gait, such as the start, mid-walk, and end, or phases like left foot lift, right foot lift, and both feet on the ground. Each cluster is treated as a "bag of instances", furthermore, we introduce an attention mechanism that assigns greater importance to instances (frames) within each bags that are more likely to exhibit scoliosis-related abnormalities. This enables more effective aggregation of information along the temporal dimension.
Subsequently, we employ another attention mechanism to assign greater importance to the bags that are more likely to exhibit scoliosis-related abnormalities, thereby addressing the inherent variability in gait sequences. The combination of MIL and attention allows the model to prioritize the most informative instances (frames) for scoliosis detection, filtering out irrelevant information and reducing the risk of overfitting to unnecessary details. At the same time, this approach makes our method particularly robust to noise and variations in gait sequences, which are common in real-world data collection.

In summary, this paper contributes from three key aspects: 
(1) We introduce Gait-MIL, the first framework that applies Multiple Instance Learning (MIL) to scoliosis detection using gait as a biomarker. This novel approach enables a refined focus on key segments within gait sequences, which are particularly relevant for identifying scoliosis indicators. 
(2) Our framework employs an attention mechanism to identify and emphasize key frames likely to display scoliosis-related abnormalities. Additionally, by integrating a clustering method, we highlight pivotal moments or clusters within the sequence, thereby enhancing performance by focusing on essential temporal patterns. 
(3) To evaluate the effectiveness of the proposed Gait-MIL , we conduct extensive experiments on the first large-scale dataset specifically designed for video-based scoliosis classification. The results strongly validate our Gait-MIL’s effectiveness, particularly in enhancing the discrimination of Neutral cases, which are often challenging due to their subtle indicators. 

\section{Methodology}
\label{sec:Methodology}

The overview of our proposed Gait-MIL is demonstrated in Fig~\ref{fig:framework}, and the technical details are introduced as follows.

\subsection{Frame-Level Clustering}
\label{Sampling and Clustering}
In the task of scoliosis classification, each subject \( X_i \) contains a sequence of gait frames \( \{ f_{i1}, f_{i2}, \dots, f_{iT} \} \), where \( i \) denotes the subject index and \( T \) represents the length of the complete gait sequence (number of frames). However, during training, the sampler randomly selects \( S \) frames for each subject in a batch to from all available frames of each subject to form the training sequence, i.e., \( \{ f_{ij} \}_{j=1}^{S} \), where \( S \leq T \). This sampling introduces complexity and dynamic variability into the sequence, making it impractical to use these sequences as instances directly since the gait frames no longer retain their original temporal order.
To address this, K-Means~\cite{macqueen1967some}is introduced to cluster the frames within the sequence into multiple bags (phases) . This clustering method effectively handles the heterogeneity of different phases, allowing the model to extract phase-specific features from the gait data. Therefore, the sampled gait frames \(  f_{ij}  \) are partitioned into multiple bags (phases), i.e., \( \{ b_{i1}, b_{i2}, \dots, b_{iK} \} \), where \( K \) represents the number of bags formed after clustering. 
This clustering approach enables the model to extract representative features from different gait phases. 
\vspace{-10 pt}

\subsection{Attention-Based Pooling}
More specifically, even after clustering the frames into different phases, the importance of individual frames varies. To effectively handle the variability and importance of individual frames within each phase, we introduce an attention-based pooling layer that operates on the temporal dimension (i.e., frame dimension) within each phase \( b_{ik} \). Using this attention mechanism, we perform a weighted aggregation of the frames \( f_{ij} \) within each phase to identify the most salient frames that reflect scoliosis-related features, thereby generating the phase-level feature representation \( h_{ik} \).

\subsection{Bag-Level Feature Aggregation}
As described above, we cluster the frames within the sequence of each subject into multiple bags (phases) \( B_i = \{ b_{i1}, b_{i2}, \dots, b_{iK} \} \). For each instance, we apply attention-based pooling to obtain the aggregated feature \( h_{iK} \). Subsequently, an attention mechanism is used to assign different weights \( \alpha_{iK} \) to each bag \( b_{iK} \), aggregating the bag features \( h_{iK} \). The global feature representation is then obtained as:~\(H_i = \sum_{k=1}^{K} \alpha_{ik} h_{iK}
\)

\subsection{Gait-Based Scoliosis Classification}
Inspired by previous work~\cite{zhou2024gait}, we recognize gait as a biomarker for scoliosis and introduce Gait-MIL, aimed at efficiently automating the scoliosis classification process.

As illustrated in Fig~\ref{fig:framework}, the input feature map is denoted as \( \mathbb{R}^{N \times 1 \times S \times H \times W} \), where \( N \), \( 1 \), \( S \), \( H \) and \( W \) represents the batch size, the number of channels, the number of frames, height and width, respectively. Subsequently, K-Means clustering is employed to partition the \( S \) frames into distinct bags, each representing different temporal phases of the gait sequence. This results in the following: \( \mathbb{R}_1^{N \times 1 \times S_1 \times H \times W} \), \( \mathbb{R}_2^{N \times 1 \times S_2 \times H \times W} \),\ \dots, \, \( \mathbb{R}_K^{N \times 1 \times S_K \times H \times W} \), where \( S_1 + S_2 + \dots + S_K = S \), \(K\) represents  the number of bags (phases). After completing the clustering, we apply a ResNet-based backbone network \( F \) for feature extraction and representation, yielding feature maps:~\( f_i = F(R_i) \in \mathbb{R}^{N \times c \times s_i \times H \times W} \), where \( i = 1, 2, \dots, K \). 

Next, we apply Attention-Based Pooling (AP) to perform weighted aggregation on \( f_i \), identifying the key frames that best reflect scoliosis, and subsequently generating the feature representation \( h_i \) for this phase. The function is formulated by:~\(h_i = AP(f_i) \in \mathbb{R}^{N \times c \times H \times W} \), where \( i = 1, 2, \dots, K \). 

Immediately following, an attention mechanism is used to aggregate bag features into a global representation, allowing the model to allocate more attention to the most representative bags, thereby emphasizing key features, the function is formulated by: ~\(H = ATT(h_1,h_2,\dots,h_K) \in \mathbb{R}^{N \times c \times H \times W} \).

Horizontal Pooling (HP) further divides these feature maps and aggregates them into vectors \( v_s \) through global pooling, performing global pooling across 16 horizontal segments to achieve detailed feature learning and representation. The function is formulated by:~\( z = Maxpool(v_s) + Avgpool(v_s) \).

We further use a separate fully connected layer to map them into the metric space. Finally, we employ the widely-used BNNeck~\cite{luo2019bag} to adjust feature space, and the triplet and cross-entropy losses are utilized to drive the training process, such as:

\(
L_{\text{Triplet}} = \frac{1}{N_{\text{valid}}} \sum_{\substack{a,p,n \\ y_a = y_p \neq y_n}} \max(m + d(a,p) - d(a,n), 0),
\)
where  \(N_{valid}\)  denotes the number of triplets with non-zero loss in one batch.  \(a, p and n \) represent anchor, positive, and negative samples, respectively.  \(d(a, p)\)  and  \(d(a, n)\)  denote the distances between the anchor-positive and anchor-negative pairs, respectively. The function is formulated by:~\(L_{\text{ce}} = - \sum_{i=1}^{n} y_i \log(\hat{y}_i).
\)
The overall loss function is formulated by:~\(
L_{\text{Total}} = L_{\text{Triplet}}+L_{\text{ce}}.
\)

\section{Experiments}
\label{sec: Experiments}

\subsection{Dataset}
Scoliosis1K~\cite{zhou2024gait} is the first large-scale video-based dataset for scoliosis classification to date. The subjects are divided into three categories: positive (Cobb angle greater than 10 degree), neutral (Cobb angle approximately 10 degree), and negative (Cobb angle less than 10 degree).
The dataset comprises 1,050 adolescent volunteers from middle schools in China, containing a total of 447,900 sequence frames, to protect privacy, all images are presented in silhouette format. The videos are recorded at 720p with participants walking along a corridor, and the camera positioned 1.4 to 4.2 meters away. Each sequence has around 300 frames at 15 fps. Scoliosis experts classified participants through visual assessments and the Adams test, without watching the videos. The videos are labeled based on their assessments. The detailed information is shown in Table ~\ref{table:scoliosis1k}.

The dataset is divided into 745 sequences for the training set and 748 sequences for the test set. We strictly follow the evaluation protocol set by the dataset owners, in the training set, the ratio of positive:neutral:negative samples is maintained at 1:1:8, with the sequence counts being 74, 74, and 596, respectively. 
\vspace{-0.2 cm}



\begin{table}[ht]
\centering
\caption{Summary statistics of the Scoliosis1K dataset.}
\resizebox{0.49\textwidth}{!}{ 
\begin{tabular}{lllll}
\hline
\textbf{Attributes}    & \textbf{\#All} & \textbf{\#Positive} & \textbf{\#Neutral} & \textbf{\#Negative}\\ \hline

\textbf{Participants} & 1050         & 176              & 82               & 792               \\ 
\textbf{Sequences}    & 1493         & 493              & 200              & 800               \\ 
\textbf{Sex (F/M)}              & 641/409      & 113/63           & 49/33            & 479/313           \\  \hline
\textbf{Attributes}    & \textbf{All} & \textbf{Positive} & \textbf{Neutral} & \textbf{Negative}\\ \hline

\textbf{Age (years)}& 15.2 ± 1.5   & 14.3 ± 1.0       & 14.0 ± 0.6       & 15.5 ± 1.5        \\ 
\textbf{Height (cm}& 163.2 ± 8.8  & 161.6 ± 7.1      & 161.4 ± 6.7      & 163.7 ± 9.3       \\ 
\textbf{Weight (kg)}& 51.9 ± 10.7  & 48.3 ± 8.4       & 46.7 ± 7.8       & 53.3 ± 11.1       \\ 
\hline
\end{tabular}
}
\vspace{-0.3 cm}
\flushleft\footnotesize \textit{Note: Values of Age, Height and Weight are presented as mean ± standard deviation (std).}

\label{table:scoliosis1k}

\end{table}

\vspace{-0.6 cm}
\subsection{Evaluation Protocol}
\label{sec:Protocol}
In this work, the model is evaluated using three commonly used metrics in recognition tasks: Accuracy, Sensitivity, and Specificity, which are defined
as follows:
\begin{itemize}
    \item \textbf{Accuracy}: The ratio of correctly classified samples to the total number of samples.
    \item \textbf{Sensitivity}: The ratio of true positives (correctly identified scoliosis cases) to the total number of actual positive cases.
    \item \textbf{Specificity}: The ratio of true negatives (correctly identified normal cases) to the total number of actual negative cases.
\end{itemize}

\subsection{Implementation Details}

Our Gait-MIL model is developed based on the OpenGait~\cite{fan2023opengait}. The input resolution of dataset is set to 64 × 44 pixels. The anchor and the positive come from the same subject but consist of different frames. We set the margin(i.e., m) for the triplet loss to 0.2. The number of sampled frames \(S\) and instances  \(K\) is set to 30 and 3, respectively. The other settings are consistent with those of ScoNet-MT~\cite{zhou2024gait}.

\section{Results}
\vspace{-5 pt}
\subsection{Comparison with State-of-the-Art}
\vspace{-5 pt}
We compare our Gait-MIL with the current state-of-the-art deep learning-based method ScoNet-MT~\cite{zhou2024gait}, as well as some traditional scoliosis detection techniques.

\begin{figure}[h]  
    \centering
    \includegraphics[width=1\linewidth]{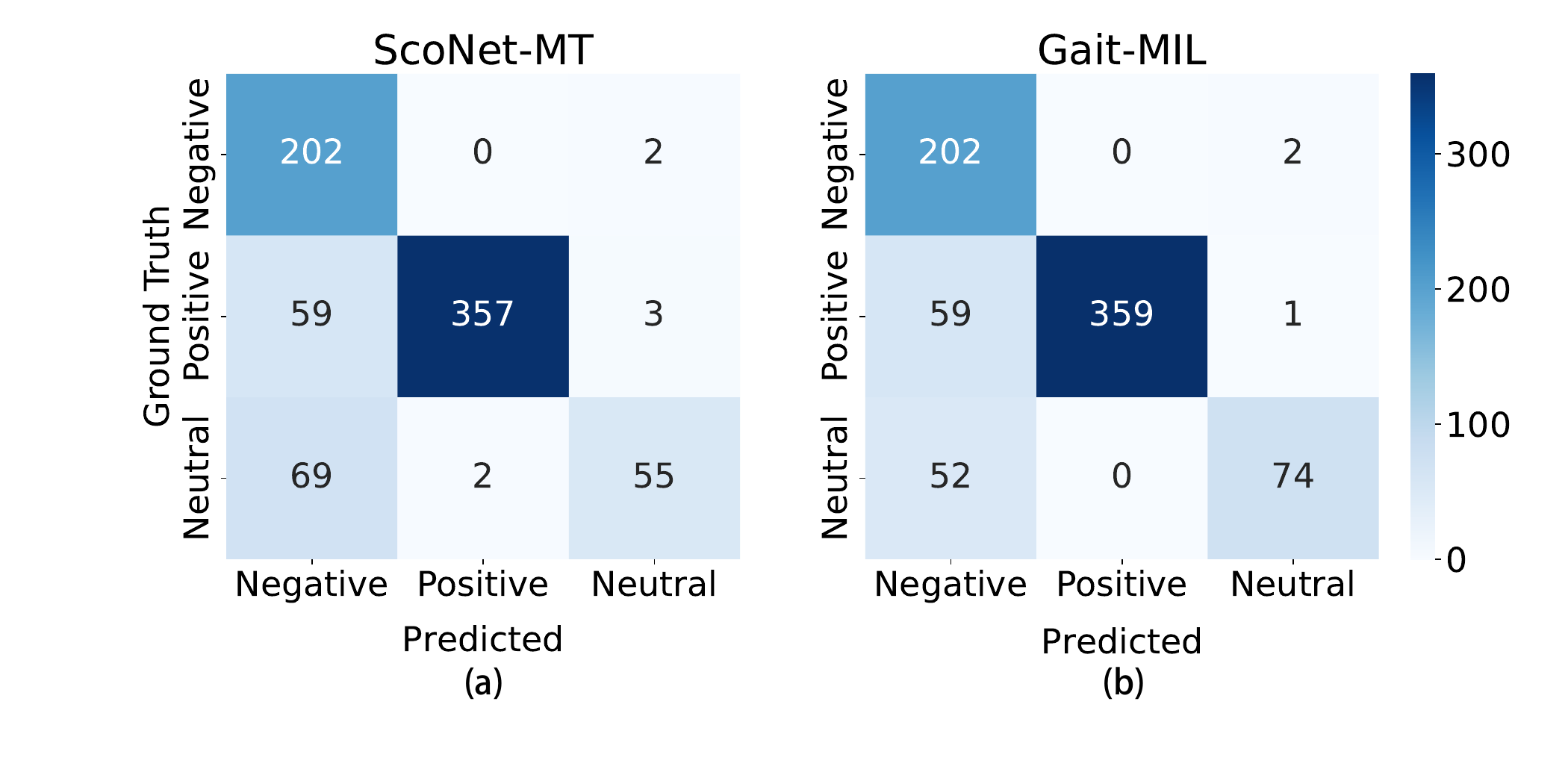}  
    \caption{Comparison of Confusion Matrices for Gait-MIL and ScoNet-MT.}
    \label{fig:linear_con}
\vspace{-0.4cm}
\end{figure}

Gait-MIL demonstrates notable improvements over traditional methods and prior deep learning models, particularly in balancing sensitivity and specificity. 
It surpasses ScoNet and ScoNet-MT with an accuracy of 84.8\%.
While ScoNet excels in sensitivity (100\%), it significantly lags in specificity (33.2\%), leading to many false positives. Gait-MIL maintains a high sensitivity of 99.0\% while increasing specificity to 79.6\%, showcasing its effectiveness in reducing misclassifications. This balance highlights the strength of Gait-MIL’s attention mechanism and multi-instance learning in accurately distinguishing between scoliosis classifications. As shown in Fig~\ref{fig:linear_con}, the most challenging aspect lies in accurately classifying Neutral cases. The state-of-the-art method (ScoNet-MT) often misclassifies Neutral cases as Negative, which could lead to misdiagnoses. In contrast, our Gait-MIL demonstrates improved classification for Neutral cases~\cite{janicki2007scoliosis,goldberg2008scoliosis}, effectively identifying some potentially ambiguous cases and reducing the risk of error diagnosis. The more exciting is Gait-MIL demonstrates high sensitivity for positive cases, with no samples being misclassified as having scoliosis.
\vspace{0 cm}
\begin{table}[h]  
    \centering
     \caption{Performance comparison of different methods.}
    \begin{tabular}{llll}
    \hline
    \textbf{Method}    & \textbf{Accuracy} & \textbf{Sensitivity} & \textbf{Specificity} \\ \hline
    \multicolumn{4}{c}{Traditional methods} \\ \hline
    Adams Test \cite{karachalios1999ten} & -                & 84.4\%              & \textbf{95.2\%}     \\ 
    Scoliometer \cite{karachalios1999ten} & -                & 90.6\%              & 79.8\%              \\
    \hline
    \multicolumn{4}{c}{Deep learning-based methods} \\ \hline
    ScoNet \cite{zhou2024gait}     & 51.3\%           & \textbf{100.0\%}     & 33.2\%              \\ 
        ScoNet-MT \cite{zhou2024gait}  & 82.0\%  & 99.0\%               & 76.5\%              \\
    Gait-MIL (ours)  & \textbf{84.8\%}  & 99.0\%               & 79.6\%              \\
    \hline
    \end{tabular}
   
    \label{table:performance}
    \vspace{-9 pt}
\end{table}

\subsection{Ablation Studies}

\textbf{Influence of MIL}. To evaluate the influence of MIL on model performance, we design an ablation experiment that compares the Gait-MIL with and without the MIL. As shown in Table~\ref{table:ablation_study1}, the performance of the Gait-MIL model improves after incorporating Multiple Instance Learning (MIL). Specifically, the accuracy increases from 82.2\% to 84.8\%, and the specificity improves from 77.8\% to 79.6\%. Although the sensitivity remains relatively similar with and without MIL, at 99.5\% and 99.0\%, respectively, the increase in specificity indicates that MIL helps the model better identify negative samples (normal cases), reducing false positives.

\begin{table}[ht]
\centering
\caption{Ablative results with and without MIL.}
\resizebox{0.45\textwidth}{!}{ 
\begin{tabular}{llll}
\hline
\textbf{Method}    & \textbf{Accuracy} & \textbf{Sensitivity} & \textbf{Specificity} \\ \hline

Gait-MIL (w/o MIL)  & 82.2\%  & \textbf{99.5}\%               & 77.8\%              \\
Gait-MIL  & \textbf{84.8\%}  & 99.0\%             & \textbf{79.6\%}              \\
\hline
\end{tabular}
}
\vspace{-0.3 cm}
\label{table:ablation_study1}
\end{table}

\textbf{Influence of class imbalance}. To explore the effects of class imbalance in real-world, the evaluation protocol includes additional configurations with varying positive, neutral, and negative sample ratios. Specifically, the sequence counts are 186:186:373 for the 1:1:2 ratio, 124:124:497 for the 1:1:4 ratio, and the 1:1:16 ratio. The results are shown in Table ~\ref{tab:class imbalance.}. These settings reflect real-world scenarios where negative samples are predominant, allowing for a comprehensive evaluation of the model’s performance under varying degrees of class imbalance. The results clearly demonstrate that Gait-MIL exhibits great stability in handling class imbalance problems, particularly maintaining high accuracy even on highly imbalanced datasets. This indicates that Gait-MIL exhibits excellent adaptability and robustness in dealing with real-world imbalanced data.

\begin{table}[ht]
\centering
\caption{Ablative results of class imbalance.}
\begin{tabular}{l|lll}
\hline
\textbf{Pos:Neu:Neg} & \textbf{ScoNet} & \textbf{ScoNet-MT} &\textbf{Gait-MIL} \\ \hline
1:1:2   & 91.4\%  & 95.2\% & \textbf{97.8\%} \\
1:1:4   & 88.6\%  & 90.5\%  & \textbf{92.5\%} \\
1:1:8   & 51.3\%  & 82.0\%  & \textbf{84.8\%}\\
1:1:16  & 23.7\%   & 49.5\%  &  \textbf{53.2\%}\\
 \hline
\end{tabular}

\label{tab:class imbalance.}
\end{table}

\vspace{-0.6 cm}

\section{Conclusion}
\label{sec:Conclusion}
In this study, we propose Gait-MIL, an Attention-Guided Deep Multi-Instance Learning method that effectively detects scoliosis through gait patterns. Extensive experimental results demonstrate our Gait-MIL significantly improves detection accuracy, particularly in enhancing the discrimination of Neutral cases, which are often challenging due to their subtle indicators. Our proposed method is efficient and non-invasive, enabling large-scale screening and supporting early prevention and treatment in resource-limited areas.

\section{ACKNOWLEDGMENTS}
This work was partially supported by US National Science Foundation IIS-2412195,  CCF-2400785 and the Cancer Prevention and Research Institute of Texas (CPRIT) award (RP230363). The other authors declare no conflict of interest.

\section{Compliance with Ethical Standards}
This research study was conducted retrospectively using human subject data made available in open access. Ethical approval was *not* required as confirmed by the license attached with the open-access data. This study was performed in line with the principles of the Declaration of Helsinki.


\bibliographystyle{IEEEbib}
\bibliography{refs}
\end{document}